\documentclass[sigplan]{acmart}

\usepackage{graphicx} 
\usepackage{balance}  
\usepackage{microtype} 
\usepackage{booktabs} 
\usepackage{multirow} 
\usepackage{placeins}
\usepackage{enumitem}
\usepackage{verbatim}
\usepackage{tikz}
\usetikzlibrary{arrows.meta,positioning,fit,calc,backgrounds}
\definecolor{LHBlue}{HTML}{0072B2}
\definecolor{LHOrange}{HTML}{E69F00}
\definecolor{LHGreen}{HTML}{009E73}
\definecolor{LHGray}{HTML}{4D4D4D}
\definecolor{LHLite}{HTML}{F2F2F2}

\AtBeginDocument{%
  \providecommand\BibTeX{{%
    \normalfont B\kern-0.5em{\scshape i\kern-0.25em b}\kern-0.8em\TeX}}}

\setcopyright{none}

\acmConference[Agent Skills '26]{Agent Skills Workshop at ACM CAIS}{May 26, 2026}{San Jose, CA}

\begin{document}

\title{LH-Bench: Skill-Grounded Evaluation of Long-Horizon Agents on Subjective Enterprise Tasks}

\author{Ishan Gupta}
\authornote{Equal contribution.}
\affiliation{%
  \institution{Independent}
  \country{USA}
}
\email{345ishaan@gmail.com}

\author{Abhishek Chandwani}
\authornotemark[1]
\affiliation{%
  \institution{Metaphi Inc.}
  \country{USA}
}
\email{abhishek@metaphi.ai}

\begin{abstract}
Binary success metrics work when a task has a single correct answer. They fail for long-horizon enterprise work, where agents must coordinate tools over dozens of steps, produce multiple intermediate artifacts, and satisfy subjective process constraints such as design-system discipline, source grounding, and safe iterative editing. In this setting, evaluation requires procedural knowledge, not just an outcome checker.

We present \textbf{LH-Bench}, a benchmark and evaluation design in which expert-authored \texttt{SKILL.md} artifacts serve as the bridge between execution and evaluation. Skills encode workflow expectations as observable rubric boundaries, while curated artifact contracts and human preference judgments provide independent validation. We instantiate LH-Bench in two environments: \textbf{Figma-to-code} (33 real \texttt{.fig} tasks against the Figma API via MCP) and \textbf{Programmatic content} (183 chapters across 41 courses), and evaluate three autonomous agent harness families end-to-end.

Expert-authored skills make LLM judges materially more reliable than LLM-authored rubrics alone ($\kappa{=}0.60$ vs.\ $0.46$ on the same runs), and independent human preferences recover the same primary ranking boundary ($p{<}0.05$). Skill-level decomposition exposes agent trade-offs that aggregate artifact scores hide, and structured verifier feedback enables recovery from 70.3\% of observed execution errors. We release the benchmark artifacts, rubrics, and a source-grounded human reasoning dataset spanning subject matter expert annotations, chapter plans, and pairwise preferences.
\end{abstract}



\maketitle


\section{Introduction}
The strongest existing agent benchmarks are still dominated by tasks with crisp verifiers: unit tests, exact answers, or binary environment success. Enterprise work is different. Agents must maintain state over long horizons, reason over multiple artifacts, and satisfy workflow constraints that are real but not directly executable as tests. A front-end implementation can compile yet violate the design system; a generated tutorial can be fluent yet ungrounded in the provided sources. For these tasks, binary correctness is both too weak to rank systems reliably and too coarse to diagnose failure.

This is fundamentally a \emph{skills} problem. What makes a long-horizon run good is not just the final artifact, but whether the agent executes the right procedure: inspect before editing, extract tokens before components, verify before shipping, cite before claiming. Those expectations are procedural knowledge. If they remain implicit, evaluation becomes ad hoc and hard to scale. If they are encoded explicitly as expert-authored skills, the same artifact can guide execution and define observable criteria for judging execution afterward.

We therefore frame \textbf{LH-Bench} as a skill-grounded evaluation design for subjective enterprise work. In LH-Bench, expert-authored \texttt{SKILL.md} artifacts specify workflow phases, failure modes, and completion criteria. We pair these skills with curated artifact contracts and human preference judgments, yielding a three-pillar design that can score both process quality and output quality. We instantiate LH-Bench in two environments:
\begin{itemize}
    \item \textbf{Figma-to-code:} 33 real \texttt{.fig} tasks in which agents navigate the Figma API via MCP, extract structure and assets, and iteratively build front-end implementations.
    \item \textbf{Programmatic content:} 183 chapters across 41 courses in which agents work inside curated data rooms to create code-driven videos, animations, and presentations with source-grounded citations.
\end{itemize}

These environments sit squarely in the long-horizon regime identified by prior capability studies \cite{kwa2025metr}: success depends on setup, decomposition, recovery, and context management over many interdependent steps. They also stress capabilities outside the verifiable settings where recent reasoning post-training has been most successful, especially mathematics and coding \cite{su2025rewardbridge}.

Our contributions are:
\begin{enumerate}
    \item \textbf{A skill-grounded evaluation design} for subjective long-horizon work, combining expert-authored skills, artifact contracts, and human preference validation.
    \item \textbf{Evidence that expert-authored skills improve judge reliability:} on the same 92 Figma-to-code runs, expert-authored rubric skills raise mean pairwise agreement from $\kappa{=}0.46$ to $0.60$.
    \item \textbf{A two-environment benchmark} that evaluates three autonomous harness families end-to-end and exposes harness-level differences invisible to binary pass/fail scores.
    \item \textbf{A recovery analysis} showing that agents recover from 70.3\% of observed errors when verifier hooks return structured feedback, with recoverability driven more by feedback quality than by raw error frequency.
    \item \textbf{An open release} of benchmark artifacts, rubrics, and a human reasoning dataset containing subject matter expert (SME) annotations, chapter plans, cited source spans, and pairwise preference data.
\end{enumerate}

\section{Related Work}

\paragraph{Binary agent benchmarks are insufficient for subjective enterprise work.}
WebArena \cite{zhou2024webarena}, OSWorld \cite{xie2024osworld}, SWE-bench \cite{jimenez2024swebench}, AgentBench \cite{liu2024agentbench}, and GAIA \cite{mialon2023gaia} established interactive evaluation for LLM agents, but their dominant feedback signal is still binary success or unit-test completion. Even more demanding settings such as WorkArena \cite{drouin2024workarena} and SWE-Bench Pro \cite{deng2025swebenchpro} remain primarily completion- or unit-test-graded. LH-Bench differs in making subjective process quality first-class: it evaluates whether the agent executed the right workflow, not only whether it produced a superficially acceptable end state.

\paragraph{Our environments extend prior design-to-code and multi-tool settings.}
Design2Code \cite{si2024design2code}, FronTalk \cite{wu2025frontalk}, FullFront \cite{sun2025fullfront}, and FrontendBench \cite{zhu2025frontendbench} study front-end generation, but largely from screenshots or text requirements. LH-Bench instead starts from real Figma artifacts, requiring API-based inspection, asset extraction, token discovery, and regression-safe iteration. Likewise, MINT \cite{wang2024mint} and ToolLLM \cite{qin2024toolllm} study multi-turn tool use, but not the long-horizon, multi-artifact workflows where verification and repair dominate.

\paragraph{We build on long-horizon evaluation and LLM judging, but contribute the missing procedural layer.}
Prior work shows that reliable task completion degrades sharply with horizon length \cite{kwa2025metr,ultrahorizon2025}, that sparse final rewards are inadequate for long episodes \cite{anonymous2025reinforcing}, and that conversational long-horizon evaluation in API settings poses distinct control challenges \cite{barres2025tau2bench}. In parallel, LLM-as-a-Judge work demonstrates strong alignment with human preferences but also exposes sensitivity to prompt framing and rubric quality; multi-agent judging improves robustness but still leaves the rubric-design problem open \cite{zheng2023judging,shi2024judging,li2024llmasjudgesurvey,chen2025multiagent}. Our contribution is not confidence calibration, judge ensembling, or judge selection. It is a skill-grounded rubric design for subjective enterprise trajectories, and an empirical demonstration that \emph{expert-authored skills with observable boundaries} make judging long-horizon work materially more reliable.

\paragraph{LH-Bench is an Agent Skills benchmark.}
SkillsBench \cite{li2025skillsbench} shows that skill-level decomposition reveals failure modes hidden by aggregate scores, ResearchRubrics \cite{sharma2025researchrubrics} shows that rubric design matters for ranking reliability, and DeepResearch Bench II \cite{li2026deepresearchbench2} shows the value of expert-derived fine-grained rubrics for long-form report evaluation. LH-Bench contributes a different object: long-horizon agent \emph{trajectories} in enterprise environments. Its novelty is not expert authorship alone, but treating \texttt{SKILL.md} as a dual-use artifact that both guides execution in-loop and defines observable rubric boundaries for post-hoc evaluation, exposing harness-level differences in orchestration, compaction, and recovery.

\begin{table}[t]
  \caption{Positioning of LH-Bench relative to existing agent benchmarks. Existing benchmarks rely on binary or unit-test evaluation; LH-Bench introduces skill-grounded, multi-tier scoring.}
  \label{tab:positioning}
  \centering
  \resizebox{\columnwidth}{!}{%
  \begin{tabular}{lccccc}
    \toprule
    \textbf{Benchmark} & \textbf{Tasks} & \textbf{Multi-turn} & \textbf{Real env.} & \textbf{Expert rubrics} & \textbf{Artifact eval} \\
    \midrule
    WebArena \cite{zhou2024webarena} & 812 & \texttimes & Web & \texttimes & Binary \\
    VisualWebArena \cite{koh2024visualwebarena} & 910 & \texttimes & Web & \texttimes & Binary \\
    SWE-bench \cite{jimenez2024swebench} & 2294 & \texttimes & Code & \texttimes & Unit tests \\
    OSWorld \cite{xie2024osworld} & 369 & \texttimes & OS & \texttimes & Binary \\
    $\tau$-bench \cite{yao2025taubench} & 200 & \checkmark & API & \texttimes & Binary \\
    Design2Code \cite{si2024design2code} & 484 & \texttimes & \texttimes & \texttimes & VLM \\
    MINT \cite{wang2024mint} & varies & \checkmark & \texttimes & \texttimes & Binary \\
    \textbf{LH-Bench (ours)} & \textbf{216} & \checkmark & \checkmark & \checkmark & \textbf{Multi-tier} \\
    \bottomrule
  \end{tabular}%
  }
\end{table}
\FloatBarrier

\section{Skills as Execution and Evaluation Artifacts}
\label{sec:harness}
\subsection{Why skills are the core artifact}
LH-Bench evaluates end-to-end \emph{agent harnesses}, not base models in isolation. Across Claude Code, Codex CLI, and Gemini CLI, the shared substrate is now familiar: sandboxed shell execution, file-system tools, MCP-based extensibility, and long interactive trajectories. What differs in practice is not tool availability but orchestration: when the agent inspects, when it commits to an implementation plan, when it verifies, and how it recovers. Those decisions are exactly what expert-authored skills encode.

In LH-Bench, each environment ships with a \texttt{SKILL.md} artifact that plays a dual role. \emph{At execution time}, it specifies the recommended workflow, common failure modes, and non-negotiable constraints. \emph{At evaluation time}, the same workflow phases define rubric boundaries that are observable from the transcript. For example, ``token file created before components'' is a verifiable event, not a vague notion of good planning. This dual use is the key design choice: skills are not auxiliary documentation, but the mechanism that makes subjective process quality inspectable at scale.

\subsection{Autonomous execution and context management}
All harnesses run autonomously inside sandboxes and interact with task environments through controlled APIs such as design extraction, browser automation, source extraction, rendering, and deployment. Tasks are stateful and routinely exceed context limits, so harnesses must compact prior interaction history while preserving commitments, open TODOs, and verified facts. We therefore evaluate the full harness stack, including compaction policy, retry behavior, and repair strategy, rather than a single model call.

\subsection{Verifier hooks and recovery signals}
\label{sec:verification}
Verification is a first-class interface rather than a final post-hoc check. Build failures, preview errors, rubric violations, and visual mismatches are surfaced as structured machine-readable feedback that the agent can act on immediately. In Figma-to-code, for example, a preview call triggers automated checks for runtime exceptions, blank renders, and broken routes; the returned payload includes both a diagnosis and the next required action. This turns verification into an in-loop repair signal instead of an offline score.

\paragraph{Verification hook availability.}
Preview verification hooks are implemented as post-tool-use hooks for Claude Code (via the Agent SDK hook interface) and as after-tool subprocess hooks for Gemini CLI (via \texttt{.gemini/hooks/}). Codex CLI does not expose an equivalent native hook mechanism; it can invoke the \texttt{create\_app\_preview} MCP tool but receives raw tool output without automatic post-processing.

\subsection{Extensible Tool Interface}
All three CLIs support tool extensibility via MCP (Model Context Protocol) servers or equivalent custom tool definitions. LH-Bench exposes environment-specific capabilities---Figma structure inspection, asset export, source extraction, scene generation, previewing, and deployment---through standalone tools that every harness can call. This decouples benchmark logic from harness internals and keeps the comparison focused on reasoning and orchestration.

\section{Benchmark Environments}
\label{sec:environments}
LH-Bench currently contains two enterprise environments. Both are tool-rich, stateful, and evaluated with identical tool access plus environment-specific expert skills. The environments were selected because they force agents to coordinate planning, retrieval, editing, verification, and repair over long horizons, while still admitting concrete artifact contracts.

\subsection{Figma-to-Code Environment}
In Figma-to-code, agents take real Figma artifacts as input and iteratively produce front-end implementations across 33 \texttt{.fig} tasks. The key challenge is that the agent cannot treat the task as screenshot copying: it must inspect nested structure, discover routes, extract reusable tokens and assets, and preserve prior work across repeated edits.

\paragraph{Action space.}
Agents interact through (i) Figma MCP calls for design structure, styles, components, and asset export, (ii) file and shell tools for implementation, (iii) a preview tool that catches build and runtime failures, and (iv) a deployment tool that publishes static builds for evaluation and reproducible inspection.

\paragraph{Workflow constraints.}
The expert skill enforces workflow discipline: inspect before coding, implement all frames rather than cherry-picking, avoid interactive scaffolding, use deployment-safe asset paths, and checkpoint through preview before major changes. These constraints are intentionally transcript-visible so they can become rubric boundaries at evaluation time.

\paragraph{Ground truth and verification.}
Each task includes a \texttt{ground\_truth/} directory with a \texttt{manifest.json} (frame metadata: name, node ID, viewport, target route) and 2x PNG exports per frame. The evaluated build is exercised through Playwright to capture screenshots, and a VLM judge compares those screenshots against ground truth frame by frame. We also run programmatic checks for successful builds, deployment accessibility, and component coverage. Pure infrastructure failures are excluded from model comparisons.

\paragraph{Task complexity axes.}
\label{sec:complexity_axes}
We curate tasks along four complexity axes to ensure the benchmark spans a wide capability range: (i)~\emph{application category}---e-commerce, SaaS dashboards, portfolio sites, landing pages---each imposing different design-system conventions; (ii)~\emph{frame count} (ranging from single-frame layouts to designs with 70+ frames), which determines the scope of navigation and cross-page consistency the agent must maintain; (iii)~\emph{image and asset density}, from icon-light text layouts to media-heavy product grids requiring bulk asset export and format selection; and (iv)~\emph{route navigation complexity}, from single-page designs to multi-route applications with nested navigation hierarchies and responsive breakpoints. This deliberate stratification ensures the benchmark tests agents across the full difficulty spectrum rather than clustering at a single complexity level.

\subsection{Programmatic Content Environment}
Programmatic content captures a different but equally important enterprise workflow: given a curated data room, the agent must produce code-driven media that explains complex concepts faithfully for a particular audience. The outputs are not raw video generations but \emph{programs} that render to media: Remotion videos with narration, Manim animations, or React/Framer Motion presentations. We evaluate 183 chapters across 41 courses.

\paragraph{Action space.}
Agents use three tool layers: (1)~\emph{source tools} that normalize heterogeneous documents into line-numbered markdown with a structured index, enabling line-level citation; (2)~\emph{generation tools} that create scenes, animations, or slides with synchronized audio; and (3)~\emph{verification tools} that check rendering, synchronization, and viewport constraints.

\paragraph{Multi-turn conversation simulation.}
Tasks simulate iterative expert interaction: the agent produces a sequence of chapters over a persistent workspace, then receives revision requests such as ``compare GRPO with PPO'' or ``add a gradient-flow visual.'' Strong performance therefore requires \emph{scene coherence} across turns and \emph{fast editing}: targeted changes without wholesale regeneration.

\paragraph{Grounding and verification.}
Source extraction produces a persistent structured index with per-source line counts and section headings, enabling judges to verify source grounding and penalize unsupported claims. We detail the verifier hooks here because they are central to recovery analysis: before compilation, generated React or animation code is parsed for structural errors; during rendering, Remotion compilation failures are surfaced directly to the agent; after TTS generation, audio--video duration synchronization is checked at render time; and presentation outputs are validated with screenshot and viewport checks. These verifier hooks make programmatic content amenable to the same recovery-style analysis as Figma-to-code, even though the artifacts are multimedia rather than webpages.

\begin{table}[t]
  \caption{Representative LH-Bench tasks, included in the main text to illustrate why binary pass/fail misses the procedural burden of these environments.}
  \label{tab:task_examples}
  \centering
  \small
  \resizebox{\columnwidth}{!}{%
  \begin{tabular}{p{2.4cm}p{4.2cm}p{6.2cm}}
    \toprule
    \textbf{Environment} & \textbf{Representative task} & \textbf{Why it is hard} \\
    \midrule
    Figma-to-code & Multi-route product or SaaS interface from a real \texttt{.fig} file & The agent must inspect nested frames, extract assets and tokens, implement all routes, and preserve previously-correct pages during iterative edits. \\
    Programmatic content & Chapter revision such as ``compare GRPO with PPO'' or ``add a gradient-flow visual'' & The agent must retrieve supporting evidence from multiple sources, maintain prior visual style, edit only the affected scenes, and keep narration synchronized with updated visuals. \\
    \bottomrule
  \end{tabular}%
  }
\end{table}
\FloatBarrier

\section{Benchmark Construction}
\label{sec:construction}
We construct LH-Bench from expert workflows and enterprise-representative artifacts. Every task includes (i) environment state, such as a \texttt{.fig} file or a data room of sources; (ii) expert-authored skills and rubrics; and (iii) verifiers over intermediate and final artifacts.

\subsection{Subject matter expert artifacts}
For each environment, subject matter experts (SMEs) author the workflow artifacts that drive both execution and evaluation. In Figma-to-code, four experienced front-end engineers author \texttt{SKILL.md} guidance plus the process rubrics used to score trajectories. In programmatic content, instructional-design SMEs author chapter plans, design notes, cited source spans, and the rubric criteria used for evaluation. The crucial design principle is that rubric boundaries are \emph{observable}: they refer to events that can be verified from the transcript or generated artifacts, rather than to vague notions of quality.

\subsection{SME annotation workflows}
\label{sec:sme_annotation}

\paragraph{Figma-to-code dataset curation.}
Ground truth is built from an automated pipeline with expert curation. We source candidate designs from Figma Community and enterprise design partners, filter for enterprise complexity rather than template simplicity, duplicate them into evaluation accounts, and extract structure through the production Figma API. Expert selection balances the four complexity axes from Section~\ref{sec:complexity_axes}.

\paragraph{Programmatic content source-grounded annotation.}
To build faithful ground truth for programmatic content, we provide SMEs with an annotation interface that supports chapter-level task specification and fine-grained source grounding. The interface segments sources into line-addressable spans and uses a highlight-to-cite interaction to attach exact evidence to each chapter, producing high-quality citations with explicit metadata.

\subsection{Artifact contracts and structured traces}
\label{sec:artifact_contract}
LH-Bench makes subjective tasks scorable by requiring agents to emit \emph{interim artifacts}---for example per-frame ground-truth images, manifests, chapter citation targets, and renderable scene programs---that serve as deterministic hooks for downstream verification (schemas in Appendix~\ref{app:gt_schema}). We also preserve structured traces aligned to workflow phases, so judges consume a decision-relevant action sequence rather than an unbounded raw transcript.

\subsection{Versioned infrastructure and parallel judging}
\label{sec:infra}
Task definitions, rubrics, and skill versions are tracked independently. The pipeline (Appendix~\ref{app:pipeline}) exposes execution, evaluation, and leaderboard endpoints; agents build artifacts in ephemeral sandboxes and upload final outputs to object storage. Evaluation runs three judges in parallel (Table~\ref{tab:judges}): a trajectory judge for planning and recovery, a process judge for skill execution, and an output judge for artifact fidelity. Judges emit structured JSON grades per tier, enabling both leaderboard ranking and fine-grained diagnosis.

\subsection{Open release}
We release more than prompts and final scores. The LH-Bench release includes task artifacts, rubric versions, environment-specific skills, verifier configurations, and a human reasoning dataset. For programmatic content this dataset includes SME chapter decompositions, global design notes, line-level citation spans, and pairwise preference outcomes; for Figma-to-code it includes expert rubric definitions and preference judgments over deployed artifacts. The intended use is twofold: reproducible evaluation of long-horizon agents and supervision data for future work on skill induction, rubric learning, and post-training for enterprise tasks.

\paragraph{Public artifacts.} All datasets are released on HuggingFace:
\begin{itemize}\itemsep0pt
  \item Figma-to-code tasks: \url{https://huggingface.co/datasets/metaphilabs/frontend-figma-to-code}
  \item Figma-to-code expert preferences: \url{https://huggingface.co/datasets/metaphilabs/figma2code-expert-preferences}
  \item Programmatic content tasks: \url{https://huggingface.co/datasets/metaphilabs/remotion-video-gen}
  \item Programmatic content expert preferences: \url{https://huggingface.co/datasets/metaphilabs/remotion-video-preferences}
\end{itemize}

\begin{table}[t]
  \caption{LH-Bench judges and the artifacts they consume.}
  \label{tab:judges}
  \centering
  \small
  \begin{tabular}{lll}
    \toprule
    \textbf{Judge} & \textbf{Inputs} & \textbf{Scores} \\
    \midrule
    Trajectory & Transcript + tool traces & Planning, recovery \\
    Process    & Transcript + skill rubric & Workflow compliance \\
    Output     & Ground truth + screenshots & Visual fidelity \\
    \bottomrule
  \end{tabular}
\end{table}
\FloatBarrier

\section{Evaluation Framework}
Evaluation uses two primary tiers, each scored on anchored 1--5 scales and reported separately to preserve diagnostic value.

\paragraph{Process tier (skill score).}
Three LLM judges (Gemini 3.1 Pro, Claude Sonnet 4.6, GPT-5.2) score structured traces and transcripts on four expert-authored process rubrics with observable boundaries aligned to workflow phases: design inspection, token extraction, component architecture, and build verification. We report means and cross-judge variance (Section~\ref{sec:rubric_analysis}).

\paragraph{Output tier (artifact score).}
A VLM judge (Gemini 3) compares Playwright-captured screenshots against expert ground-truth images frame by frame on eight rubrics covering visual fidelity, layout accuracy, and interaction correctness (Appendix~\ref{app:output_rubrics}).

\paragraph{Scoring.}
\label{sec:scoring}
Each tier is a weighted average ($S_{\mathrm{tier}} = \sum_i w_i s_i$, $\sum w_i = 1$). Output scores provide an aggregate quality signal; process scores preserve the skill-level detail needed for diagnosis, ablation, and future post-training.

\section{Experiments}

\subsection{Experimental Setup}
We evaluate three agent harness families---Claude Code, Codex CLI, and Gemini CLI---across seven configurations (harness $\times$ model; Appendix~\ref{app:harness}) on two environments: Figma-to-code (33 tasks) and programmatic content (183 chapters across 41 courses). Each configuration executes autonomously with identical tool access and the same environment-specific expert skill. Skill-tier evaluation uses one flagship judge model from each family. A controlled ablation compares matched Figma-to-code runs with and without \texttt{SKILL.md} (Section~\ref{sec:ablation}).

For programmatic content, SMEs provide both reasoning artifacts and scoring criteria: chapter decompositions, global design notes, source-span annotations, and a five-skill rubric covering content selection, narrative structure, visual hierarchy, information density, and source grounding. These artifacts serve as ground truth for rubric synthesis and VLM-based evaluation.

\subsection{Figma-to-Code Results}
\label{sec:results}
Table~\ref{tab:output_leaderboard} reports output scores, Table~\ref{tab:skill_leaderboard} reports skill scores, and Table~\ref{tab:rubric_breakdown} decomposes the skill score by rubric. All three primary candidates include 95\% bootstrap confidence intervals (1{,}000 resamples over tasks).

\begin{table}[t]
  \caption{Figma-to-code output scores (VLM judge, 1--5 scale). Seven model configurations across three harness families. 95\% bootstrap CI reported for primary candidates.}
  \label{tab:output_leaderboard}
  \centering
  \small
  \resizebox{\columnwidth}{!}{%
  \begin{tabular}{llcc}
    \toprule
    \textbf{Rank} & \textbf{Agent harness / model} & \textbf{Score} & \textit{n} \\
    \midrule
    1 & Codex (GPT-5.2 Pro) & 4.27 & 18 \\
    2 & Claude Code (Opus 4.6) & $4.19 \pm 0.28$ & 32 \\
    3 & Codex (GPT-5.2) & $3.94 \pm 0.36$ & 31 \\
    4 & Claude Code (Opus 4.5) & 3.88 & 29 \\
    5 & Gemini CLI (Gemini 3.1 Pro) & $3.73 \pm 0.44$ & 29 \\
    6 & Claude Code (Sonnet 4.5) & 3.66 & 22 \\
    7 & Gemini CLI (Gemini 3 Pro) & 3.59 & 22 \\
    \bottomrule
  \end{tabular}%
  }
\end{table}

\begin{table}[t]
  \caption{Figma-to-code skill scores (3 LLM judges, 1--5 scale). 95\% bootstrap CI over tasks. Per-judge columns: Gemini 3.1 Pro, Claude Sonnet 4.6, GPT-5.2. Var.\ is population variance across judges.}
  \label{tab:skill_leaderboard}
  \centering
  \small
  \resizebox{\columnwidth}{!}{%
  \begin{tabular}{llccccc}
    \toprule
    \textbf{Rank} & \textbf{Agent / model} & \textbf{Score (\textit{n})} & \textbf{Gemini} & \textbf{Claude} & \textbf{GPT} & \textbf{Var.} \\
    \midrule
    1 & Claude Code (Opus 4.6) & $3.27 \pm 0.14$ (31) & 3.50 & 3.37 & 2.92 & 0.14 \\
    2 & Codex (GPT-5.2) & $3.16 \pm 0.15$ (31) & 3.51 & 3.11 & 2.82 & 0.15 \\
    3 & Gemini CLI (3.1 Pro) & $2.80 \pm 0.11$ (29) & 3.06 & 2.70 & 2.67 & 0.06 \\
    \bottomrule
  \end{tabular}%
  }
\end{table}

\begin{table}[t]
  \caption{Figma-to-code skill scores decomposed by rubric (average across 3 LLM judges, 1--5 scale). 95\% bootstrap CI on overall score. Rubric columns: Inspect.=Design Inspection, Token=Token \& Style Extraction, Comp.=Component Architecture, Build=Build Verification. Bold indicates highest per-rubric score. $\kappa$ row shows mean pairwise Cohen's kappa.}
  \label{tab:rubric_breakdown}
  \centering
  \small
  \resizebox{\columnwidth}{!}{%
  \begin{tabular}{llccccc}
    \toprule
    \textbf{Rank} & \textbf{Agent / model} & \textbf{Score (\textit{n})} & \textbf{Inspect.} & \textbf{Token} & \textbf{Comp.} & \textbf{Build} \\
    \midrule
    1 & Claude Code (Opus 4.6) & $3.27 \pm 0.14$ (31) & 3.40 & \textbf{2.88} & \textbf{3.58} & 3.21 \\
    2 & Codex (GPT-5.2) & $3.16 \pm 0.15$ (31) & \textbf{3.52} & 2.64 & 3.34 & 2.97 \\
    3 & Gemini CLI (3.1 Pro) & $2.80 \pm 0.11$ (29) & 2.95 & 1.66 & 3.28 & \textbf{3.45} \\
    \midrule
    & $\kappa$ (agreement) & 0.60 & 0.50 & 0.58 & 0.34 & 0.67 \\
    \bottomrule
  \end{tabular}%
  }
\end{table}
\FloatBarrier

\subsection{Programmatic Content Results}
\label{sec:video_results}

We evaluate programmatic content generation with a VLM-as-judge setup: a Gemini 3.1 Pro judge scores each rendered chapter against SME-authored rubrics covering content relevance, visual design, pedagogical effectiveness, audio--visual synchronization, and technical accuracy (Appendix~\ref{app:rubric_video}). Scores are normalized to $[0, 1]$ and reported as course-level means with 95\% bootstrap confidence intervals ($n{=}41$ courses, 183 chapters). We then validate these rankings against $n{=}275$ matched human SME pairwise preferences.

\begin{table}[t]
  \caption{Programmatic content: VLM artifact scores (Gemini 3.1 Pro judge, normalized 0--1). Scores are course-level means of per-chapter normalized scores with 95\% bootstrap CIs ($n{=}41$ courses, 183 chapters).}
  \label{tab:vlm_artifact_scores}
  \centering
  \small
  \begin{tabular}{lcc}
    \toprule
    \textbf{Agent / model} & \textbf{3-pt score} & \textbf{5-pt score} \\
    \midrule
    Claude Code (Opus 4.6)  & $0.588 \pm 0.093$ & $0.612 \pm 0.069$ \\
    Gemini CLI (3.1 Pro)    & $0.507 \pm 0.087$ & $0.526 \pm 0.063$ \\
    Codex (GPT-5.2)         & $0.441 \pm 0.071$ & $0.478 \pm 0.044$ \\
    \bottomrule
  \end{tabular}
\end{table}

\paragraph{Rubric granularity analysis.}
Both 3-point and 5-point scales preserve the same harness ranking (Claude Code $>$ Gemini CLI $>$ Codex), confirming that the result is not an artifact of rubric granularity. The 5-point scale yields tighter confidence intervals (38\% reduction) and reduces VLM ties by 49\%, bringing VLM tie behavior closer to human annotators (Appendix~\ref{app:human_vlm}).

\subsection{Rubric-Level Analysis}
\label{sec:rubric_analysis}
Table~\ref{tab:rubric_breakdown} reveals why the skill tier is useful: it separates bottlenecks in workflow execution from bottlenecks in output quality.

\paragraph{Design token and style extraction is a universal bottleneck.}
All three agents score lowest on \emph{token \& style extraction} (Claude Code 2.88, Codex 2.64, Gemini CLI 1.66). This is the clearest evidence that long-horizon front-end work is not merely code generation. Even strong models can implement components competently once coding begins, but they often fail to externalize the design system first. That gap then propagates into downstream fidelity errors.

\paragraph{Component and layout architecture is consistently strong.}
All agents score highest or near-highest on \emph{component \& layout architecture} (Claude Code 3.58, Codex 3.34, Gemini CLI 3.28), indicating that current models are comparatively good at structural implementation once the workflow has been set up correctly.

\paragraph{Compensatory skill profiles emerge across agents.}
Codex leads on \emph{design inspection} (3.52 vs.\ Claude Code's 3.40) but trails on \emph{build verification} (2.97 vs.\ Gemini CLI's 3.45). Gemini CLI, despite the lowest overall score, achieves the strongest build-verification behavior, suggesting a harness that is relatively persistent at repair even when upstream extraction is weak. These compensatory profiles are precisely what aggregate leaderboard scores hide.

\paragraph{Task-level difficulty gradient.}
The deliberate complexity stratification (Section~\ref{sec:complexity_axes}) produces a meaningful difficulty spread rather than a benchmark full of trivial wins. Applying the expert pass/fail threshold ($\leq$3 = fail, $\geq$4 = pass) across all rated tasks ($n{=}31$), 71\% of tasks are discriminating: at least one harness passes and at least one fails. This is where skill-level differences matter most.

\FloatBarrier
\subsection{Ablation: Removing Expert Skills}
\label{sec:ablation}

We ablate the effect of expert-authored \texttt{SKILL.md} in Figma-to-code by rerunning all three harness families \emph{without} the expert workflow (single-line task description, no procedural guidance) on three tasks spanning the difficulty gradient, then comparing against matched runs \emph{with} \texttt{SKILL.md}. Tool access is held constant. Table~\ref{tab:ablation_skill} reports per-harness skill scores ($n{=}7$ paired runs).

\begin{table}[t]
  \caption{\texttt{SKILL.md} ablation: mean skill scores (3-judge avg, 1--5) with and without expert workflow guidance across 3 tasks. $n$ = number of paired task runs per harness.}
  \label{tab:ablation_skill}
  \centering
  \small
  \resizebox{\columnwidth}{!}{%
  \begin{tabular}{llccccc}
    \toprule
    \textbf{Agent} & \textbf{Condition} & \textbf{Insp.} & \textbf{Token} & \textbf{Arch.} & \textbf{Build} & \textbf{Overall} \\
    \midrule
    Claude Code ($n{=}3$) & Without & 3.44 & 2.78 & 3.44 & 3.22 & 3.23 \\
                          & With    & 3.78 & 3.11 & 3.56 & 2.89 & 3.38 \\
                          & $\Delta$ & $+$0.34 & $+$0.33 & $+$0.12 & $-$0.33 & $+$0.15 \\
    \midrule
    Codex ($n{=}2$)       & Without & 2.17 & 2.50 & 2.34 & 1.00 & 2.06 \\
                          & With    & 3.00 & 2.84 & 3.17 & 2.67 & 2.93 \\
                          & $\Delta$ & $+$0.83 & $+$0.34 & $+$0.83 & $+$1.67 & $+$0.87 \\
    \midrule
    Gemini CLI ($n{=}2$)  & Without & 3.50 & 1.50 & 3.00 & 3.00 & 2.78 \\
                          & With    & 3.17 & 1.67 & 3.17 & 3.34 & 2.83 \\
                          & $\Delta$ & $-$0.33 & $+$0.17 & $+$0.17 & $+$0.34 & $+$0.05 \\
    \midrule
    \textbf{Pooled} ($n{=}7$) & $\Delta$ & $+$0.28 & $+$0.28 & $+$0.37 & $+$0.56 & \textbf{$+$0.33} \\
    \bottomrule
  \end{tabular}%
  }
\end{table}

Preliminary evidence from this small-scale ablation ($n{=}7$ paired runs) suggests that expert skills matter, but not uniformly across harnesses. Codex gains the most ($+$0.87 overall), with build verification improving from 1.00 to 2.67; without \texttt{SKILL.md}, it effectively fails to preview and iterate. Claude Code gains modestly ($+$0.15), with improvements concentrated in early-stage inspection and token extraction. Gemini CLI changes little ($+$0.05), suggesting that its built-in iteration pattern partly compensates for missing workflow guidance. Across harnesses, the strongest operational signal is deployment success: only 2 of 7 runs without \texttt{SKILL.md} produce a deployable artifact. We also observe a 42\% execution-cost reduction for Claude Code (mean \$6.45$\to$\$3.75) when expert skills reduce exploratory backtracking. Because the study is underpowered (2--3 runs per harness), we treat it as directional rather than definitive; Appendix~\ref{app:versioning} describes the infrastructure for scaling this analysis.

\subsection{Inter-Judge Agreement}
\label{sec:judge_agreement}
We use three judges from different model families (Gemini 3.1 Pro, Claude Sonnet 4.6, GPT-5.2) and report cross-judge variance and Cohen's $\kappa$ \cite{cohen1960kappa} as our primary agreement measures.

\paragraph{Figma-to-code.} Mean pairwise Cohen's $\kappa$ (quadratic weights) is 0.60 ($n{=}92$ runs, 360 ratings), indicating moderate-to-substantial agreement \cite{landis1977kappa}; per-rubric $\kappa$ ranges from 0.34 (component architecture) to 0.67 (build verification). All three judges preserve the same rank ordering across harnesses despite absolute-score offsets consistent with known calibration differences across LLM families \cite{li2024llmasjudgesurvey}.

\paragraph{Expert-authored vs.\ LLM-authored rubrics.}
We compare agreement under two rubric versions on the same 92 runs (Table~\ref{tab:rubric_versions}). Under v1.1 (8 LLM-authored rubrics), $\kappa = 0.46$; under v1.2 (4 expert-authored rubrics), $\kappa = 0.60$---a $+0.15$ improvement. The result is the paper's central empirical point: better skills produce better judges, even when the judge models and evaluated runs are fixed.

\begin{table}[t]
  \caption{Inter-judge agreement by rubric version (same judge families, same 92 runs). $\kappa$ is quadratic-weighted Cohen's kappa.}
  \label{tab:rubric_versions}
  \centering
  \small
  \resizebox{\columnwidth}{!}{%
  \begin{tabular}{lcc}
    \toprule
    \textbf{Metric} & \textbf{v1.1 (LLM, 8 rubrics)} & \textbf{v1.2 (Expert, 4 rubrics)} \\
    \midrule
    Mean pairwise $\kappa$ & 0.46 & \textbf{0.60} \\
    \quad Gemini--Claude & 0.44 & \textbf{0.65} \\
    \quad Gemini--GPT & 0.31 & \textbf{0.52} \\
    \quad Claude--GPT & 0.63 & \textbf{0.64} \\
    Mean variance & 0.25 & \textbf{0.10} \\
    \bottomrule
  \end{tabular}%
  }
\end{table}

\paragraph{Convergent validity across evaluation tiers.}
Three independent signals---VLM artifact fidelity, three-judge skill evaluation, and pairwise human preference (Table~\ref{tab:human_preference})---all recover the same primary ranking boundary. Because the tiers use different inputs, different judges, and different criteria, this convergence is strong evidence that expert-authored skill verifiers are useful ranking signals rather than artifacts of a single judging setup.

\paragraph{Human preference evaluation.}
To validate LLM-based rankings with direct human judgment, a domain expert performed 135 pairwise preference evaluations on Figma-to-code outputs across 31 tasks, comparing agent-built UIs side by side without access to LLM scores or agent identity (Table~\ref{tab:human_preference}). Bradley-Terry Elo rankings \cite{bradley1952rank} place Codex and Claude Code in a statistically indistinguishable top tier ($p{=}0.67$, Cohen's $h{=}0.08$), with both significantly preferred over Gemini CLI ($p{=}0.036$ and $p{=}0.047$ respectively). Position bias is absent (52\% A-rate, binomial $p{=}0.72$), and preferences are nearly transitive (4.2\% cycle rate). Winner quality ratings also differ across agents (Kruskal--Wallis $p{=}0.048$): Codex wins are rated higher on average (3.88/5) than Claude Code (3.40/5) or Gemini CLI (3.35/5), suggesting higher output peaks despite similar top-tier preference rates.

\paragraph{Expert pass/fail classification.}
The domain expert also assigns absolute quality ratings on a 5-point scale: $\leq$3 = fail, 4 = design system well-defined but assets and fine-grained spacing still require one engineering sprint, 5 = production-ready. Overall, 60\% of winning outputs pass ($\geq$4), with substantial variance across models: Codex GPT-5.2 Pro reaches 77.8\% (21/27), Codex GPT-5.2 reaches 57.9\% (33/57), Claude Code reaches 54.5\% (24/44), and Gemini CLI reaches 52.4\% (11/21). This reinforces that pairwise preference and absolute artifact quality are related but not interchangeable.

At the \emph{individual run} level, human and LLM judges show weak concordance ($\kappa{=}0.08$ output, $0.06$ skill). At the \emph{aggregate} level, both agree on the primary ranking boundary, but the LLM judges nominally separate the top two where human preferences do not, cautioning against interpreting fine-grained LLM score gaps as perceptible quality differences.

\begin{table}[t]
  \caption{Human pairwise preference (Figma-to-code): Bradley-Terry Elo from 135 votes by 1 domain expert. The top two harnesses are statistically indistinguishable ($p{=}0.67$, $h{=}0.08$); both significantly outperform Gemini CLI ($p{<}0.05$).}
  \label{tab:human_preference}
  \centering
  \small
  \begin{tabular}{lcccc}
    \toprule
    \textbf{Agent harness} & \textbf{Elo} & \textbf{95\% CI} & \textbf{Win \%} & \textit{n}\textsubscript{wins} \\
    \midrule
    Codex (GPT-5.2)        & 1054 & [1005, 1114] & 56.8 & 72 \\
    Claude Code (Opus 4.6) & 1039 & [987, 1093]  & 55.1 & 43 \\
    Gemini CLI (3.1 Pro)    &  907 & [842, 965]   & 31.1 & 21 \\
    \bottomrule
  \end{tabular}
\end{table}

\paragraph{Human--VLM alignment (programmatic content).}
We validate VLM-as-judge scoring against $n{=}275$ matched chapter-level human SME pairwise preferences (Table~\ref{tab:human_vlm_winrates}).

\begin{table}[t]
  \caption{Programmatic content: pairwise win rates from human SME preferences vs.\ VLM-derived outcomes ($n{=}275$ matched chapter-level comparisons).}
  \label{tab:human_vlm_winrates}
  \centering
  \small
  \resizebox{\columnwidth}{!}{%
  \begin{tabular}{lccc}
    \toprule
    \textbf{Agent} & \textbf{Human WR} & \textbf{VLM WR (3-pt)} & \textbf{VLM WR (5-pt)} \\
    \midrule
    Claude Code (Opus 4.6) & 81.5\% & 66.8\% & 69.0\% \\
    Codex (GPT-5.2)        & 34.2\% & 33.9\% & 33.1\% \\
    Gemini CLI (3.1 Pro)    & 34.2\% & 49.2\% & 47.8\% \\
    \bottomrule
  \end{tabular}%
  }
\end{table}

Both human and VLM judges agree that Claude Code is the top-ranked harness (human WR 81.5\%, VLM WR 69.0\%), and Codex win rates align closely (34.2\% vs.\ 33.1\%). The main divergence is at position~2: humans rate Codex and Gemini equally, while the VLM favors Gemini, likely because production polish is overweighted relative to content accuracy. Pairwise agreement improves from 46.5\% to 52.0\% under 5-point rubrics as finer granularity reduces tie asymmetry (Appendix~\ref{app:human_vlm}).

\subsection{Failure Taxonomy and Test-Time Recovery}
\label{sec:failure_taxonomy}
We extract structured error$\rightarrow$recovery events from all 96 Figma-to-code agent transcripts using an LLM-based analysis pipeline (Tables~\ref{tab:error_taxonomy},~\ref{tab:recovery_summary}).

\begin{table}[t]
  \caption{Failure taxonomy across all Figma-to-code runs ($n{=}96$). Recovery rate is the fraction of errors from which the agent successfully self-corrected.}
  \label{tab:error_taxonomy}
  \centering
  \small
  \begin{tabular}{lrrr}
    \toprule
    \textbf{Error type} & \textbf{Count} & \textbf{Recovered} & \textbf{Recovery \%} \\
    \midrule
    Tool call failure  & 419 & 278 & 66.3 \\
    Git error          &  64 &  48 & 75.0 \\
    Syntax error       &  33 &  30 & 90.9 \\
    Dependency error   &  28 &  22 & 78.6 \\
    Preview deny       &  18 &  16 & 88.9 \\
    Build error        &  11 &  11 & 100.0 \\
    Type error         &   7 &   6 & 85.7 \\
    Config error       &   6 &   1 & 16.7 \\
    Runtime error      &   4 &   3 & 75.0 \\
    \midrule
    \textbf{Total}     & \textbf{590} & \textbf{415} & \textbf{70.3} \\
    \bottomrule
  \end{tabular}
\end{table}

\begin{table}[t]
  \caption{Per-agent recovery summary (Figma-to-code). Failure rate = failed tool calls / total tool calls. Recovery rate = errors recovered / total errors.}
  \label{tab:recovery_summary}
  \centering
  \small
  \resizebox{\columnwidth}{!}{%
  \begin{tabular}{lccccccc}
    \toprule
    \textbf{Agent} & \textbf{Runs} & \textbf{Errors} & \textbf{Failure \%} & \textbf{Recovery \%} & \textbf{Preview} & \textbf{Deploy} & \textbf{Tool calls} \\
    \midrule
    Claude Code (Opus 4.6) & 32 & 152 & 7.3 & 71.1 & 30/32 & 27/32 & 2692 \\
    Codex (GPT-5.2)        & 34 & 273 & 9.4 & 74.0 & 5/34  & 21/34 & 3339 \\
    Gemini CLI (3.1 Pro)    & 30 & 165 & 8.4 & 63.6 & 30/30 & 30/30 & 2092 \\
    \bottomrule
  \end{tabular}%
  }
\end{table}

\FloatBarrier

\paragraph{Tool call failures dominate; error message quality determines recoverability.}
Tool call failures account for 71\% of all errors (419/590), including MCP timeouts, file-not-found errors, and permission denials (Figure~\ref{fig:error_landscape}). The clearest pattern is not which errors occur, but which errors are actionable: structured compiler-style feedback (syntax, type, build) yields $>$85\% recovery, whereas ambiguous configuration signals yield only 17\%.

\paragraph{Agents recover from 70\% of errors, with distinct correction strategies.}
Across 590 errors, agents self-correct 70.3\% of the time (Table~\ref{tab:recovery_summary}), showing that verifier hooks are not only evaluators but useful runtime skills in their own right. The harnesses adopt distinct repair styles: Claude Code is \emph{proactive}, previewing early and often; Codex is \emph{reactive}, encountering more errors but recovering from many of them; Gemini CLI is \emph{persistent}, driving consistently to deployment.

\section{Conclusion}
The core claim of this paper is simple: evaluating long-horizon enterprise agents requires explicit procedural knowledge, and expert-authored skills are a practical way to provide it. In LH-Bench, skills do double duty. They guide the agent at execution time, and they define observable boundaries for judging the resulting trajectory. That design makes subjective enterprise work scorable without collapsing everything into a brittle binary metric.

Across Figma-to-code and programmatic content, expert-authored skills improve judge reliability, human preferences confirm the main ranking boundary, and structured verifier feedback supports real recovery during execution. Just as importantly, skill-level decomposition reveals which parts of the workflow are failing. That is the signal needed for benchmark design, harness engineering, and post-training alike.

\section{Limitations and Future Work}
(1)~Our findings are limited to two environments, albeit two that stress very different skills; extending the framework to additional enterprise domains is necessary before making claims of broad external validity. (2)~We evaluate concrete commercial harnesses rather than a model-agnostic open implementation, so model capability and harness design remain partially entangled. (3)~The \texttt{SKILL.md} ablation is underpowered and should be read directionally. A natural next step is to scale this study and investigate automated skill induction from successful and failed trajectories. (4)~Human preference validation remains expensive; improving the faithfulness of multimodal judges, especially on content accuracy versus surface polish, is still an open problem.

\bibliographystyle{ACM-Reference-Format}
\bibliography{sample-base}

\clearpage
\onecolumn
\appendix

\section*{Appendix}
\addcontentsline{toc}{section}{Appendix}

\makeatletter
\newcommand{\appentry}[2]{%
  \noindent\hyperref[#2]{\textbf{#1}}\dotfill\pageref{#2}\par\medskip}
\makeatother

\appentry{A\quad Pipeline Diagram}{app:pipeline}
\appentry{B\quad SME Annotation Tool}{app:annotation_tool}
\appentry{C\quad Ground Truth Schema Examples}{app:gt_schema}
\appentry{D\quad Harness Specifications}{app:harness}
\appentry{E\quad Skill Rubric Definitions}{app:rubrics}
\appentry{F\quad Rubric Version Comparison (v1.1 vs v1.2)}{app:rubric_versions}
\appentry{G\quad Output Tier Rubric Weights}{app:output_rubrics}
\appentry{H\quad Recovery Analysis Figures}{app:recovery_figures}
\appentry{I\quad Programmatic Content Evaluation Rubrics}{app:rubric_video}
\appentry{J\quad Human--VLM Alignment Details}{app:human_vlm}
\appentry{K\quad Preference Arena}{app:preference_arena}
\appentry{L\quad Task-Level Human Baseline}{app:task_baseline}
\appentry{M\quad Experiment Versioning Infrastructure}{app:versioning}

\vfill


\section{Pipeline Diagram}
\label{app:pipeline}

\begin{figure}[h]
  \centering
  \begin{tikzpicture}[
    font=\footnotesize,
    arrow/.style={-Latex, line width=0.8pt, draw=LHGray},
    box/.style={draw=LHGray, rounded corners=2pt, align=center, inner sep=4pt,
                minimum height=10mm, text width=28mm, fill=white},
    store/.style={box, fill=blue!5, draw=LHBlue},
    api/.style={box, fill=LHLite},
    compute/.style={box, fill=orange!6, draw=LHOrange},
    eval/.style={box, fill=green!6, draw=LHGreen},
    rowlbl/.style={font=\scriptsize\bfseries\color{LHGray}, anchor=east},
  ]
    \node[store] (tasks) {\textbf{Task dataset}\\[-1pt]{\scriptsize (HuggingFace)}\\[-1pt]{\tiny row\_id, figma\_key, prompt}};
    \node[api, right=14mm of tasks] (execute) {\textbf{/execute}\\[-1pt]{\scriptsize Dispatch agent run}};
    \node[compute, right=14mm of execute] (sandbox) {\textbf{Agent sandbox}\\[-1pt]{\scriptsize Build, preview,}\\[-1pt]{\scriptsize structured feedback}};
    \node[store, right=14mm of sandbox] (rundb) {\textbf{Run DB + artifacts}\\[-1pt]{\scriptsize Deployed URLs, media,}\\[-1pt]{\scriptsize scores, judge variance}};
    \node[store, below=18mm of tasks] (rubrics) {\textbf{Rubric dataset}\\[-1pt]{\scriptsize (HuggingFace)}\\[-1pt]{\tiny skill, anchors, weights}};
    \node[api, below=18mm of execute] (evaluate) {\textbf{/evaluate}\\[-1pt]{\scriptsize Load artifacts + rubrics}\\[-1pt]{\scriptsize Run judges in parallel}};
    \node[eval, below=18mm of sandbox] (judges) {\textbf{3 Judges (parallel)}\\[-1pt]{\scriptsize Trajectory / Process /}\\[-1pt]{\scriptsize Output}};
    \node[api, below=18mm of rundb] (leader) {\textbf{/leaderboard}\\[-1pt]{\scriptsize Aggregate + rank}\\[-1pt]{\scriptsize Confidence reporting}};
    \node[rowlbl] at ($(tasks.west)+(-3mm,0)$) {Execution};
    \node[rowlbl] at ($(rubrics.west)+(-3mm,0)$) {Evaluation};
    \draw[arrow] (tasks) -- (execute);
    \draw[arrow] (execute) -- (sandbox);
    \draw[arrow] (sandbox) -- (rundb);
    \draw[arrow] (rubrics) -- (evaluate);
    \draw[arrow] (evaluate) -- (judges);
    \draw[arrow] (judges) -- (leader);
    \draw[arrow, LHBlue!70] (rundb.south) -- ++(0,-5mm) -| (evaluate.north);
  \end{tikzpicture}
  \caption{LH-Bench execution and evaluation pipeline. Tasks and rubrics are versioned independently in HuggingFace. Agent runs produce persistent artifacts, which are graded by three judges in parallel; results flow into leaderboards.}
  \label{fig:pipeline}
\end{figure}
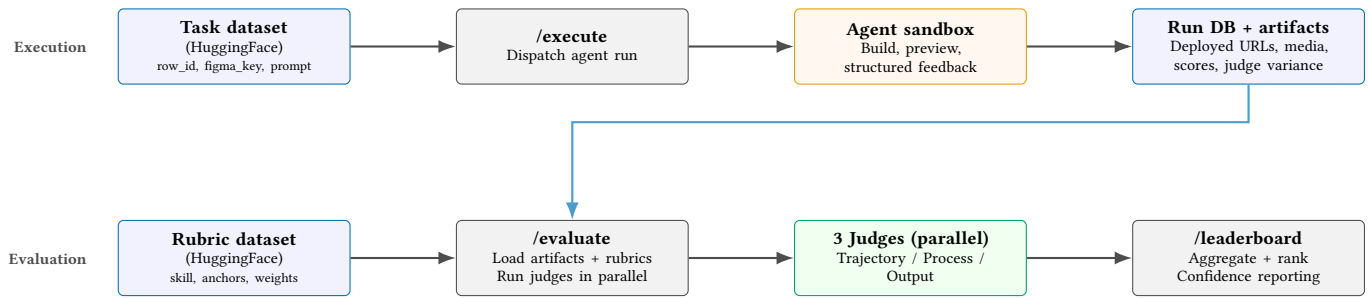


\section{SME Annotation Tool}
\label{app:annotation_tool}

\begin{figure}[h]
  \centering
  \includegraphics[width=\textwidth]{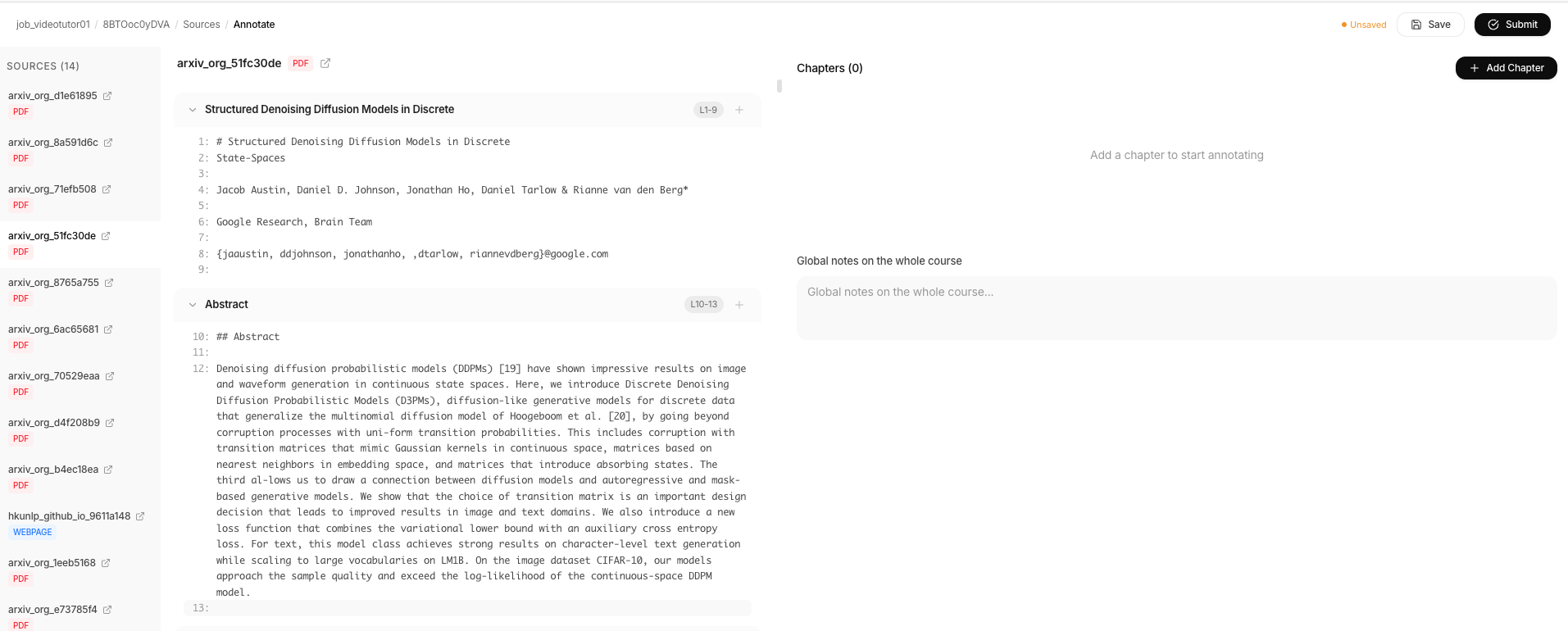}
  \caption{SME annotation interface for programmatic content. Left: source documents in the data room. Center: line-numbered document viewer with collapsible sections and highlight-to-cite interaction. Right: chapter definitions with attached source spans and global design notes.}
  \label{fig:annotation_tool}
\end{figure}

Figure~\ref{fig:annotation_tool} shows the SME annotation interface used for programmatic content ground-truth construction. The tool presents three coordinated panels:

\paragraph{Source panel (left).} Lists all documents in the task's data room (e.g., arXiv PDFs, web pages) with type badges. SMEs click a source to load it in the center panel.

\paragraph{Document viewer (center).} Renders the selected source as line-numbered markdown. Each section is collapsible and displays its line range (e.g., \texttt{L1--9}, \texttt{L10--13}). SMEs can select text spans inline using a ``highlight-to-cite'' interaction: selecting lines attaches the span (with source ID and line numbers) to the active chapter, producing structured citations of the form \texttt{source\_id:start\_line--end\_line}.

\paragraph{Chapter panel (right).} SMEs define chapters---the units of content the agent must produce---and attach source spans to each chapter. Each chapter includes a title, ordering, and the set of cited spans from the source panel. A ``Global notes'' field captures high-level design reasoning (e.g., narrative arc, emphasis priorities) that applies across all chapters.

This three-panel design enables SMEs to build granular, source-grounded annotations efficiently: the annotator reads a source, highlights relevant passages, and assigns them to chapters in a single workflow, rather than context-switching between separate tools. The resulting annotations power rubric synthesis for VLM-based artifact scoring (Section~\ref{sec:video_results}) and encode the 5-skill design rubric (content selection, narrative structure, visual hierarchy, information density, source grounding).

\section{Ground Truth Schema Examples}
\label{app:gt_schema}

\paragraph{Figma-to-code manifest} (see Section~\ref{sec:artifact_contract}):

\begin{verbatim}
{
  "figma_file_key": "oSNDllo...8YMSlD",
  "total_frames": 5,
  "frames": [
    { "name": "PDP",
      "node_id": "2176:167104",
      "gt_image": "2176-167104.png",
      "target_route": "/pdp" },
    { "name": "PLP_Category",
      "node_id": "2176:167875",
      "gt_image": "2176-167875.png",
      "target_route": "/" }
  ]
}
\end{verbatim}

\paragraph{Programmatic content annotation} (see Section~\ref{sec:artifact_contract}):

\begin{verbatim}
{
  "task_id": "video-tutor-042",
  "context_gt": [
    { "source_id": "arxiv-2301.00234",
      "start_line": 42, "end_line": 58,
      "quote": "Attention is computed as..." }
  ],
  "rubrics": [
    { "criterion": "source_grounding",
      "scale": "0/1/2",
      "anchor_2": "All claims cited" }
  ]
}
\end{verbatim}


\section{Harness Specifications}
\label{app:harness}
This appendix provides formal descriptions of the agent harnesses, tool interface, skill injection mechanism, and session recovery protocol used in LH-Bench.

\subsection{Harness Descriptions}
We evaluate three commercial agent harnesses. Each tightly couples specific models with proprietary agent logic, representing real-world deployment conditions:
\begin{itemize}
    \item \textbf{Claude Code} (Anthropic): Anthropic's coding agent with native Skill integration. Uses the Claude Agent SDK with subprocess CLI execution, MCP server support via \texttt{McpStdioServerConfig}, automatic context compaction, and session resumption. Operates in permission-bypass mode (\texttt{acceptEdits}) for autonomous execution.
    \item \textbf{Codex CLI} (OpenAI): OpenAI's lightweight coding agent. MCP servers are registered via a global registry (\texttt{codex mcp add}). Runs in \texttt{dangerously-bypass-approvals-and-sandbox} mode for full autonomy. Outputs structured JSONL event streams with transcript archival.
    \item \textbf{Gemini CLI} (Google): Google's open-source terminal agent. MCP servers are auto-discovered from a per-project \texttt{.gemini/settings.json} configuration. Uses Vertex AI authentication with wrapper scripts for environment variable isolation. Runs in trust mode (\texttt{--yolo}) for auto-approval.
\end{itemize}

\paragraph{Model family consideration.}
Claude models have been trained with awareness of the Agent Skills specification, which may confer advantages when processing Skill-formatted instructions. Codex and Gemini harnesses receive equivalent skill content but through their respective native mechanisms.

\subsection{Tool Interface}
All agents interact with environments through the Model Context Protocol (MCP). Each MCP server exposes domain-specific tools following a standardized request/response interface:

\paragraph{Figma-to-code tool categories.}
\begin{itemize}
    \item \textbf{Design extraction} (Figma MCP): \texttt{get\_figma\_file}, \texttt{get\_node\_tree}, \texttt{export\_node\_images}---retrieve design structure, component hierarchy, and rasterized assets.
    \item \textbf{File and shell} (built-in): Read, Write, Edit, Glob, Grep, Bash---standard file manipulation and command execution within the sandbox.
    \item \textbf{Preview verification} (App Preview MCP): \texttt{create\_app\_preview}, \texttt{get\_preview\_status}---build the agent's code in an ephemeral container and return structured error diagnostics (runtime exceptions, blank-page detection) or a live URL.
    \item \textbf{Deployment} (GCS MCP): \texttt{upload\_dist\_to\_gcs}---publish finalized static builds for artifact evaluation and human inspection.
    \item \textbf{Browser automation} (Playwright MCP): \texttt{browser\_navigate}, \texttt{browser\_take\_screenshot}---exercise deployed UIs and capture frame-level screenshots for visual verification.
\end{itemize}

\subsection{Skill Injection Mechanism}
\texttt{SKILL.md} files encode expert-authored procedural knowledge as structured Markdown with YAML frontmatter:

{\small
\begin{verbatim}
---
name: figma-to-code
description: Convert Figma designs to
  production-ready frontend code.
---
# Step 0: Check manifest for prior progress
# Step 1: Extract design structure via Figma MCP
# Step 2: Export assets and ground truth frames
...
\end{verbatim}
}

\paragraph{Per-harness loading.}
Each harness discovers and loads Skills using its native mechanism:
\begin{itemize}
    \item \textbf{Claude Code}: \texttt{setting\_sources=["project"]} triggers scanning of \texttt{.claude/skills/*/SKILL.md} in the project root.
    \item \textbf{Gemini CLI}: Skills are placed in \texttt{.gemini/skills/*/SKILL.md} and loaded via file-based configuration in \texttt{settings.json}.
    \item \textbf{Codex CLI}: Skill content is inlined into the system prompt, as Codex lacks a native skill-discovery mechanism.
\end{itemize}

\subsection{Manifest-Based Session Recovery}
Agents maintain a flat \texttt{manifest.json} at the project root to track execution progress:

{\small
\begin{verbatim}
{
  "preview_url": "https://...",
  "deployed_url": "https://...",
  "completed_steps": [
    "Step 1: Extract from Figma",
    "Step 2: Export assets"
  ],
  "updated_at": "2026-01-31T..."
}
\end{verbatim}
}

Step~0 of every \texttt{SKILL.md} requires reading the manifest to check prior progress, enabling session recovery across agent restarts without re-executing completed work. This is critical for long-horizon tasks where context limits or transient failures may interrupt a session.

\subsection{Containerization and Deployment}
Agent runs execute in sandboxed containers deployed via Modal (serverless). Each model configuration receives a dedicated persistent volume (e.g., \texttt{/figma-claude-opus}, \texttt{/figma-codex-52}, \texttt{/figma-gemini-31-pro}) for project state isolation. The container image includes Python~3.11, Node.js~20, and all agent CLI binaries. MCP tool servers run as co-processes within the same container, with credentials injected via \texttt{modal.Secret} at runtime.


\section{Skill Rubric Definitions (v2.1)}
\label{app:rubrics}

Table~\ref{tab:rubric_definitions} presents the four expert-authored process rubrics used for Figma-to-code skill evaluation (v2.1). Each rubric uses a 1--5 anchored scale with observable transcript evidence. The rubrics are designed around sequential workflow phases with binary-observable boundaries between score levels.

\begin{table}[h]
  \caption{Figma-to-code process rubrics (v2.1, expert-authored). Weight indicates contribution to the aggregate skill score. Anchors summarize the 1--5 scale boundaries.}
  \label{tab:rubric_definitions}
  \centering
  \small
  \resizebox{\textwidth}{!}{%
  \begin{tabular}{p{3.8cm}cp{5cm}p{5cm}}
    \toprule
    \textbf{Rubric} & \textbf{Wt.} & \textbf{What it measures} & \textbf{Key boundary (3\textrightarrow4)} \\
    \midrule
    Design Inspection \& Asset Extraction & 0.30 &
      Extent of Figma file inspection, asset export completeness, format correctness, multi-page navigation discovery &
      3 = all assets exported in correct formats; 4 = also organized (semantic filenames, directory structure, hierarchy inspection before coding) \\
    \addlinespace
    Design Token \& Style Extraction & 0.25 &
      Extraction and centralization of colors, typography, spacing, shadows, borders into a token/theme file &
      3 = token file covers 4+ of 6 categories, tokens referenced in code; 4 = also created before components, semantic naming, zero hardcoded leaks \\
    \addlinespace
    Component \& Layout Architecture & 0.25 &
      Component decomposition, pattern reuse, Auto Layout\textrightarrow CSS mapping, variant/state handling &
      3 = repeated patterns identified, layout correct, hover states; 4 = also planned upfront (visible in transcript), full variant coverage, props interface \\
    \addlinespace
    Build Verification \& Iteration & 0.20 &
      Build/preview execution, error diagnosis and fix iteration, visual verification against Figma design &
      3 = build compiles successfully; 4 = also opened preview AND made at least one fix based on visual inspection \\
    \bottomrule
  \end{tabular}%
  }
\end{table}

Each scale point includes concrete transcript indicators. For example, at score 5 (``Production-grade'') on Design Inspection, the agent uses WebP for photos, applies @2x scale factors, deduplicates repeated assets, converts SVGs to components, and maps the full navigation topology before coding. The full rubric specification with all 5 anchor levels per rubric is available in the repository at \texttt{verifiers/figma-to-code/process\_rubrics.json}.


\section{Rubric Version Comparison (v1.1 vs v1.2)}
\label{app:rubric_versions}

Table~\ref{tab:rubric_version_detail} compares the two rubric versions used in the inter-judge agreement analysis (Section~\ref{sec:judge_agreement}).

\paragraph{v1.1 (LLM-authored, 8 rubrics).} Generated by prompting an LLM to produce evaluation criteria for Figma-to-code agent trajectories. The rubrics cover fine-grained workflow steps with unequal weights (0.07--0.20) and use generic proficiency anchors (``Inadequate'' through ``Expert'') without binary-observable boundaries.

\paragraph{v1.2 (Expert-authored, 4 rubrics).} Designed by domain experts with Figma-to-code workflow knowledge. Each rubric maps to a sequential workflow phase, uses binary-observable boundaries between score levels (e.g., ``token file created before components'' is verifiable in the transcript), and includes specific transcript evidence patterns.

\begin{table}[h]
  \caption{Rubric version comparison. v1.1 uses 8 generic LLM-authored rubrics; v1.2 uses 4 domain-specific expert-authored rubrics with anchored scales.}
  \label{tab:rubric_version_detail}
  \centering
  \small
  \begin{tabular}{lll}
    \toprule
    & \textbf{v1.1 (LLM-authored)} & \textbf{v1.2 (Expert-authored)} \\
    \midrule
    Rubric count & 8 & 4 \\
    Scope & Figma-to-code (fine-grained) & Figma-to-code (workflow phases) \\
    Anchor style & Generic (``Inadequate''--``Expert'') & Observable (transcript evidence) \\
    Boundary type & Subjective & Binary-observable \\
    \midrule
    \multicolumn{3}{l}{\textit{v1.1 rubrics (non-equal weight):}} \\
    & \multicolumn{2}{l}{Design file inspection (0.07), Image asset extraction (0.20),} \\
    & \multicolumn{2}{l}{Icon/vector extraction (0.15), Design token discovery (0.15),} \\
    & \multicolumn{2}{l}{Component pattern recognition (0.10), Variant/state analysis (0.12),} \\
    & \multicolumn{2}{l}{Layout structure analysis (0.13), Build verification (0.08)} \\
    \midrule
    \multicolumn{3}{l}{\textit{v1.2 rubrics (weighted):}} \\
    & \multicolumn{2}{l}{Design inspection (0.30), Token extraction (0.25),} \\
    & \multicolumn{2}{l}{Component architecture (0.25), Build verification (0.20)} \\
    \midrule
    Mean pairwise $\kappa$ & 0.46 & 0.60 \\
    Mean variance & 0.25 & 0.10 \\
    \bottomrule
  \end{tabular}
\end{table}

The key design insight is that \textbf{domain-specific rubrics with binary-observable boundaries reduce scoring ambiguity}. For example, ``Did the agent create a token file before writing components?''\ is unambiguously verifiable from the transcript, whereas ``Did the agent demonstrate good planning?''\ requires subjective interpretation that varies across judges. This is reflected in the $+0.15$ kappa improvement (Table~\ref{tab:rubric_versions}).


\section{Output Tier Rubric Weights}
\label{app:output_rubrics}

Table~\ref{tab:output_rubric_weights} lists the eight artifact rubrics used for the Figma-to-code output tier (Tier~1). Each rubric is scored on a 1--5 scale by a VLM judge (Gemini 3) comparing Playwright-captured screenshots against expert ground-truth frame images.

\begin{table}[h]
  \caption{Figma-to-code output tier rubric weights.}
  \label{tab:output_rubric_weights}
  \centering
  \small
  \begin{tabular}{lcp{7cm}}
    \toprule
    \textbf{Rubric} & \textbf{Weight} & \textbf{What it measures} \\
    \midrule
    Component coverage & 0.20 & Percentage of design components rendered \\
    Layout accuracy & 0.18 & Spatial positioning/sizing and flex/grid correctness \\
    Colors accuracy & 0.14 & Palette fidelity (fills, gradients, borders) \\
    Typography accuracy & 0.12 & Font family/size/weight/line-height match \\
    Asset display & 0.10 & Images/icons/vector assets render correctly \\
    Visual fidelity & 0.10 & Overall visual similarity \\
    Responsive behavior & 0.08 & Adapts to multiple viewports \\
    Interaction fidelity & 0.08 & Hover/active/disabled states \\
    \bottomrule
  \end{tabular}
\end{table}


\section{Recovery Analysis Figures}
\label{app:recovery_figures}

\begin{figure}[h]
  \centering
  \includegraphics[width=0.7\textwidth]{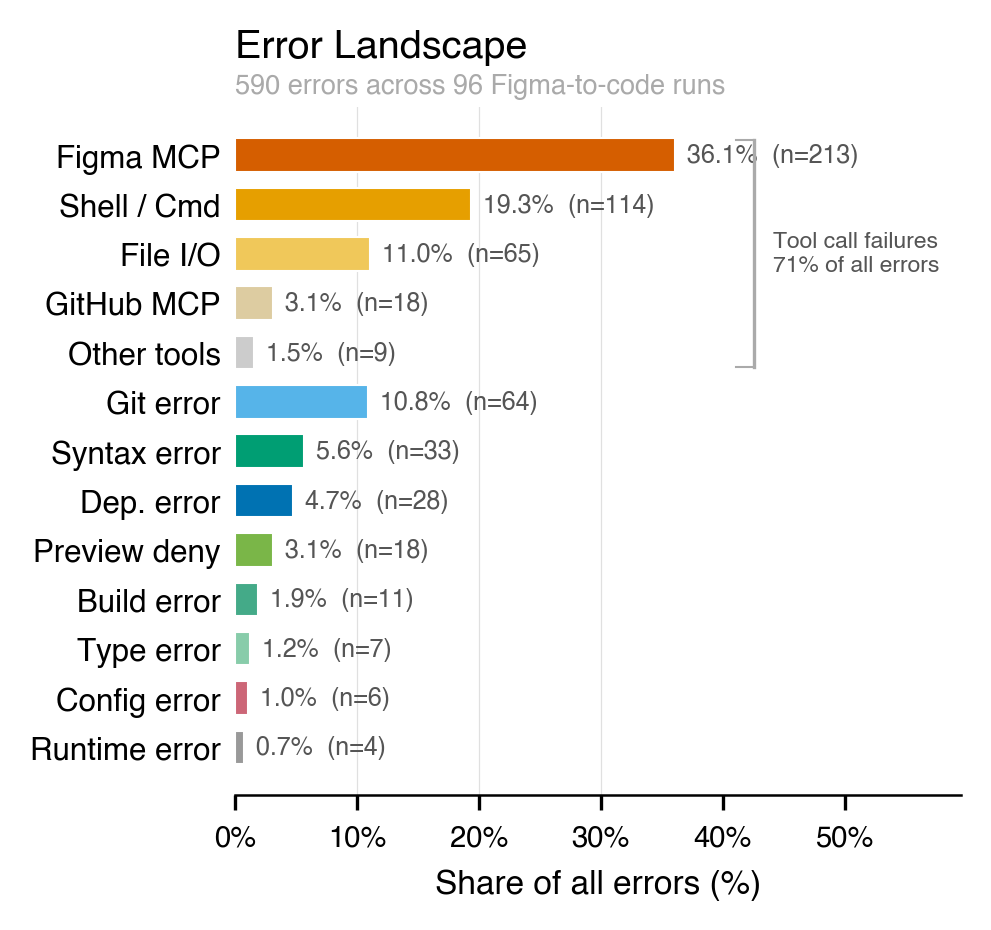}
  \caption{Error landscape across 96 Figma-to-code runs (590 total errors). Tool call failures account for 71\% of all errors; within these, Figma MCP operations are the dominant source (51\%), reflecting the difficulty of reliably invoking design-extraction APIs at scale.}
  \label{fig:error_landscape}
\end{figure}

\begin{figure}[h]
  \centering
  \includegraphics[width=0.7\textwidth]{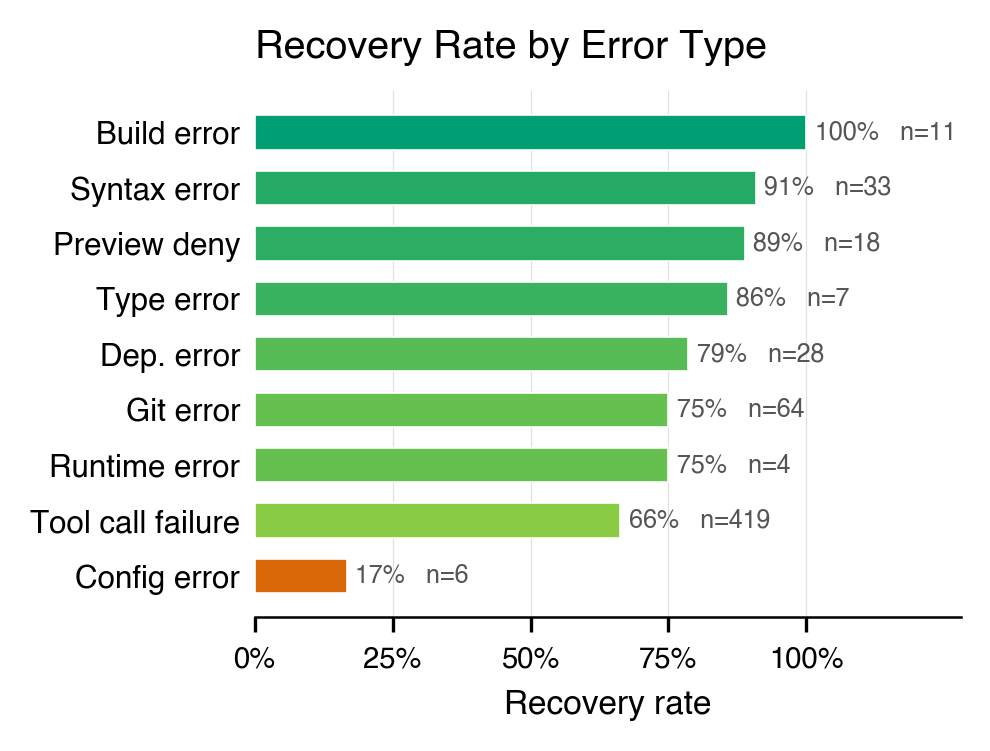}
  \caption{Recovery rates by error type. Structured compiler feedback (syntax, type, build errors) yields $>$85\% recovery; ambiguous signals (configuration errors) yield only 17\%.}
  \label{fig:recovery_by_type}
\end{figure}

\begin{figure}[h]
  \centering
  \includegraphics[width=0.7\textwidth]{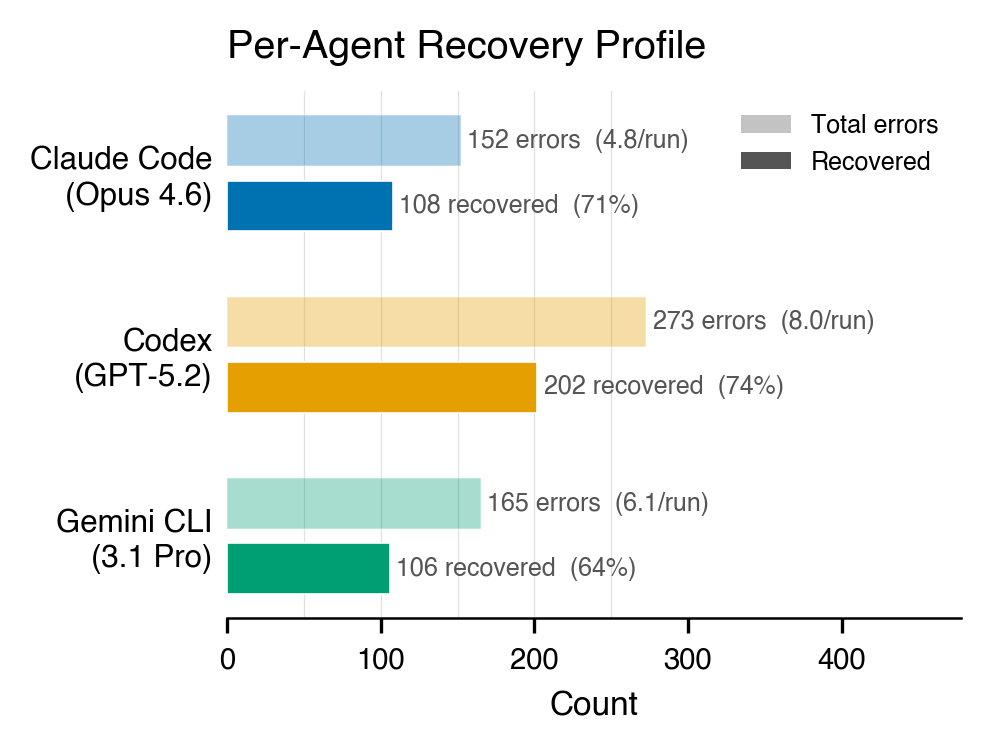}
  \caption{Per-agent recovery profiles. Codex encounters the most errors (8.0/run) yet achieves the highest recovery rate (74\%); Claude Code encounters the fewest (4.8/run) with 71\% recovery; Gemini CLI recovers 64\% with 100\% deploy completion.}
  \label{fig:agent_recovery}
\end{figure}
\FloatBarrier


\section{Programmatic Content Evaluation Rubrics}
\label{app:rubric_video}

Each chapter in the programmatic content environment is graded against five generic SME-authored rubrics plus chapter-specific criteria synthesized from annotator evaluation notes. The generic rubrics are applied uniformly across all chapters; chapter-specific rubrics are LLM-synthesized from per-chapter annotator notes to capture task-specific quality dimensions (e.g., ``correctly illustrates the attention mechanism'' for a chapter on transformers). Below we present the five generic rubrics with their full 5-point scale definitions.

\subsection{Generic Rubrics (5-point scale)}

\paragraph{1. Content Relevance and Clarity.}
Evaluates whether the video content accurately addresses the chapter instruction and presents information in a clear, logically structured manner.

\smallskip\noindent
\small
\begin{tabular}{cp{12cm}}
  \toprule
  \textbf{Score} & \textbf{Description} \\
  \midrule
  1 & Content is largely irrelevant or incoherent; fails to address the chapter instruction. \\
  2 & Addresses the topic but with major gaps, inaccuracies, or disorganized presentation. \\
  3 & Covers core points adequately; minor gaps or unclear transitions but generally on-topic. \\
  4 & Clear, well-organized content that addresses all key aspects of the instruction with minor omissions. \\
  5 & Comprehensive, precisely targeted content with logical flow; every segment directly serves the chapter objective. \\
  \bottomrule
\end{tabular}
\normalsize

\paragraph{2. Visual Design and Production Quality.}
Evaluates the aesthetic quality, consistency, and professionalism of visual elements including typography, color, layout, and animations.

\smallskip\noindent
\small
\begin{tabular}{cp{12cm}}
  \toprule
  \textbf{Score} & \textbf{Description} \\
  \midrule
  1 & Visuals are broken, missing, or unreadable; severe rendering artifacts. \\
  2 & Functional but amateurish; inconsistent styling, poor contrast, or cluttered layouts. \\
  3 & Acceptable visual quality; consistent styling with minor polish issues. \\
  4 & Professional appearance; cohesive color palette, clean typography, smooth animations. \\
  5 & Exceptional production quality; polished transitions, purposeful motion design, broadcast-quality aesthetics. \\
  \bottomrule
\end{tabular}
\normalsize

\paragraph{3. Pedagogical Effectiveness.}
Evaluates how well the video teaches the intended concept, including pacing, scaffolding, and use of examples.

\smallskip\noindent
\small
\begin{tabular}{cp{12cm}}
  \toprule
  \textbf{Score} & \textbf{Description} \\
  \midrule
  1 & No discernible teaching structure; concepts presented without context or progression. \\
  2 & Attempts to explain but lacks scaffolding; jumps between concepts without bridging. \\
  3 & Reasonable pedagogical flow; builds on prior context with adequate pacing. \\
  4 & Effective teaching with clear scaffolding, well-timed examples, and concept reinforcement. \\
  5 & Exemplary pedagogy; progressive disclosure, concrete-to-abstract scaffolding, retrieval cues, and anticipation of learner misconceptions. \\
  \bottomrule
\end{tabular}
\normalsize

\paragraph{4. Audio--Visual Synchronization.}
Evaluates alignment between narration and on-screen visuals, including timing of transitions, text highlights, and animation triggers.

\smallskip\noindent
\small
\begin{tabular}{cp{12cm}}
  \toprule
  \textbf{Score} & \textbf{Description} \\
  \midrule
  1 & Severe desynchronization; narration and visuals are unrelated or offset by multiple seconds. \\
  2 & Noticeable timing mismatches; visuals often appear before or after the relevant narration. \\
  3 & Generally synchronized; occasional minor offsets that do not impede comprehension. \\
  4 & Well-synchronized; visuals reinforce narration with consistent timing. \\
  5 & Precise synchronization; animations trigger at exactly the right narration cue, enhancing comprehension through temporal alignment. \\
  \bottomrule
\end{tabular}
\normalsize

\paragraph{5. Technical Accuracy of Visualizations.}
Evaluates the correctness of diagrams, equations, code snippets, and data representations shown in the video.

\smallskip\noindent
\small
\begin{tabular}{cp{12cm}}
  \toprule
  \textbf{Score} & \textbf{Description} \\
  \midrule
  1 & Visualizations contain fundamental errors (wrong equations, incorrect diagrams, fabricated data). \\
  2 & Partially correct but with significant errors that could mislead learners. \\
  3 & Mostly correct; minor inaccuracies that do not alter the core message. \\
  4 & Accurate visualizations with proper notation, correct relationships, and faithful data representation. \\
  5 & Technically impeccable; visualizations are precise, properly labeled, and include appropriate caveats or simplification notes where relevant. \\
  \bottomrule
\end{tabular}
\normalsize

\subsection{Chapter-Specific Rubrics}
In addition to the five generic rubrics, each chapter receives 1--3 chapter-specific rubrics synthesized by an LLM from the human annotator's evaluation criteria for that chapter. For example, a chapter on ``GRPO vs.\ PPO gradient flow'' might receive a rubric for ``Gradient diagram correctness: does the visualization accurately show the policy gradient computation path for both algorithms?'' These chapter-specific rubrics use the same scale structure (either 3-point or 5-point) as the generic rubrics and are scored by the same VLM judge.

\subsection{3-Point vs.\ 5-Point Scale Comparison}
We evaluate all rubrics under both granularities. The 3-point scale uses levels 0 (absent/incorrect), 1 (partially correct), and 2 (fully correct). The 5-point scale provides finer discrimination as shown above. Section~\ref{sec:video_results} reports aggregate results under both scales.

\subsection{Example VLM Judge Prompt (5-Point Scale)}
\label{app:vlm_prompt}

Below is an abbreviated example of the prompt sent to the VLM judge (Gemini 3.1 Pro) for a single chapter evaluation. The prompt includes: (1) a system preamble enforcing strict, evidence-based scoring; (2) course context and outline; (3) the chapter instruction; (4) the full set of rubrics (5 generic + 2 chapter-specific in this example); and (5) structured output instructions. We show the system preamble, two representative rubrics (one generic, one chapter-specific), and the output format. The full prompt for all rubrics follows the same pattern.

{\small
\begin{verbatim}
You are a strict, impartial evaluator of
AI-generated educational videos. Score only what
you observe in the video, not what it could have
been. Be critical. Reserve top scores for
genuinely exceptional work.

## Course Context
[Course description and outline omitted for space]

## Chapter Instruction
> How do Denoising Score Matching with Langevin
> Dynamics (SMLD) and DDPM learn and sample from
> complex data distributions, and why are they
> fundamentally equivalent under a score-based
> perspective?

## Rubrics

### 1. Content Relevance and Clarity
Evaluation: Does the video stay focused on
explaining how SMLD and DDPM learn/sample, and
why they are equivalent under score-based
perspective?

Scale:
  1: Mostly off-topic or incoherent; fails to
     address how SMLD/DDPM learn and sample;
     major factual errors.
  2: Touches on diffusion but with major gaps;
     equivalence missing or asserted without
     explanation.
  3: Adequate overview of SMLD and DDPM with
     minor gaps; equivalence mentioned but not
     strongly supported.
  4: Clear and well-structured; explicitly
     explains equivalence under unified
     score-based view; negligible omissions.
  5: Exceptionally focused; rigorously and
     intuitively explains SMLD-DDPM equivalence;
     consistent notation; resolves common
     confusions.

[... 4 more generic rubrics: Visual Design,
 Pedagogical Effectiveness, Audio-Visual Sync,
 Technical Accuracy ...]

### 6. Chapter Mastery: Understanding SMLD
Evaluation: How well does the video explain SMLD
as learning scores via denoising score matching
across noise levels and sampling via Langevin
dynamics?

Scale:
  1: Mentions SMLD without meaningful
     explanation; sampling missing or incorrect.
  2: High-level description but vague about noise
     levels or sampling mechanism.
  3: Explains denoising score matching and noise
     levels; basic Langevin sampling idea but
     omits key details.
  4: Clearly explains score matching across noise
     levels and annealed Langevin dynamics with
     correct update structure.
  5: Crisp end-to-end account: forward
     perturbation, score learning, principled
     sampling via annealed Langevin dynamics;
     addresses typical pitfalls.

### 7. Chapter Mastery: DDPM and Score-Based
###    Equivalence to SMLD
[Similar 5-point scale structure]

## Instructions
For each rubric:
1. Note specific evidence (timestamps, visual
   elements, narration) relevant to that rubric.
2. Match observations against each scale level.
3. Assign the integer score. Do NOT interpolate.
4. Cite specific evidence in thinking_process.

Return JSON only:
{"rubric_scores": [
  {"rubric_name": "...",
   "score": "<integer>",
   "matched_level": "<scale description>",
   "thinking_process": "Specific evidence: ..."}
]}
\end{verbatim}
}

The chapter-specific rubrics (rubrics 6--7 in this example) are synthesized from human annotator evaluation criteria for each chapter, ensuring that the VLM judge evaluates both generic production quality and chapter-specific conceptual mastery. The \texttt{design\_context} field (omitted above) additionally instructs the judge on expected visual structure---e.g., ``use parallel, side-by-side layout to emphasize SMLD--DDPM equivalence.''


\section{Human--VLM Alignment Details}
\label{app:human_vlm}

This appendix provides per-pair breakdowns of human--VLM agreement for the programmatic content environment, complementing the aggregate results in Section~\ref{sec:judge_agreement}.

\subsection{Per-Pair Agreement and Cohen's $\kappa$}

Table~\ref{tab:human_vlm_perpair} reports agreement and $\kappa$ for each agent pair under both rubric granularities. Agreement is the fraction of comparisons where the human and VLM select the same winner (or both tie); $\kappa$ is Cohen's kappa computed over the 3-class outcome (A wins, B wins, tie).

\begin{table}[h]
  \caption{Human--VLM agreement by agent pair ($n$ = matched chapter-level pairwise comparisons).}
  \label{tab:human_vlm_perpair}
  \centering
  \small
  \begin{tabular}{lccccc}
    \toprule
    \textbf{Pair} & \textit{n} & \textbf{Agree (3-pt)} & $\kappa$ \textbf{(3-pt)} & \textbf{Agree (5-pt)} & $\kappa$ \textbf{(5-pt)} \\
    \midrule
    Claude vs.\ Codex   & 92 & 57.6\% & 0.121 & 57.6\% & 0.044 \\
    Claude vs.\ Gemini  & 92 & 40.2\% & $-$0.004 & 48.9\% & 0.024 \\
    Codex vs.\ Gemini   & 91 & 41.8\% & 0.102 & 49.5\% & 0.180 \\
    \midrule
    \textbf{Overall}     & \textbf{275} & \textbf{46.5\%} & \textbf{0.073} & \textbf{52.0\%} & \textbf{0.082} \\
    \bottomrule
  \end{tabular}
\end{table}

\subsection{Tie Asymmetry and $\kappa$ Interpretation}

Under the 3-point rubric, a substantial tie asymmetry exists between human and VLM raters:

\begin{itemize}
  \item \textbf{Humans tie infrequently.} Across 275 comparisons, human SMEs produce 36 ties (13.1\%). The annotation interface supports both strict preference and explicit ties, but annotators overwhelmingly express directional preferences.
  \item \textbf{The VLM ties frequently under coarse scales.} Under the 3-point rubric, the VLM produces 74 ties (26.9\% of comparisons). Under the 5-point rubric, this drops to 38 ties (13.8\%)---a 49\% reduction that brings VLM tie behavior in line with the human tie rate (13.1\%).
\end{itemize}

Cohen's $\kappa$ is computed over a 3-class outcome space (A wins, B wins, tie). When the two raters have markedly different tie rates, $\kappa$ is deflated regardless of whether they agree on the \emph{direction} of non-tie outcomes. This explains why $\kappa$ improves from 0.073 (3-point) to 0.082 (5-point): the 5-point scale eliminates the structural tie mismatch, allowing $\kappa$ to better reflect genuine agreement on directional preferences.

\paragraph{Why win rates complement $\kappa$ here.}
The aggregate win rates (Table~\ref{tab:human_vlm_winrates}) show strong directional alignment: both human and VLM judges agree that Claude Code is the top-ranked harness (human WR 81.5\%, VLM WR 69.0\%). Codex win rates are closely aligned between human and VLM (34.2\% vs.\ 33.1\%). The primary divergence---Gemini's VLM win rate (47.8\%) exceeding its human win rate (34.2\%)---likely reflects the VLM's sensitivity to production quality dimensions where Gemini performs competitively, while human experts weight content accuracy and pedagogical effectiveness more heavily. This pattern is consistent with known biases in VLM evaluation of multimedia artifacts, where surface-level polish can inflate scores relative to content depth \cite{zheng2023judging}. The 5-point scale yields the largest agreement gains on the hardest discriminations (Codex vs.\ Gemini: $+$7.7\%, Claude vs.\ Gemini: $+$8.7\%), confirming that finer rubric granularity is most valuable precisely where evaluation is most challenging.


\section{Preference Arena}
\label{app:preference_arena}

\begin{figure}[h]
  \centering
  \includegraphics[width=\textwidth]{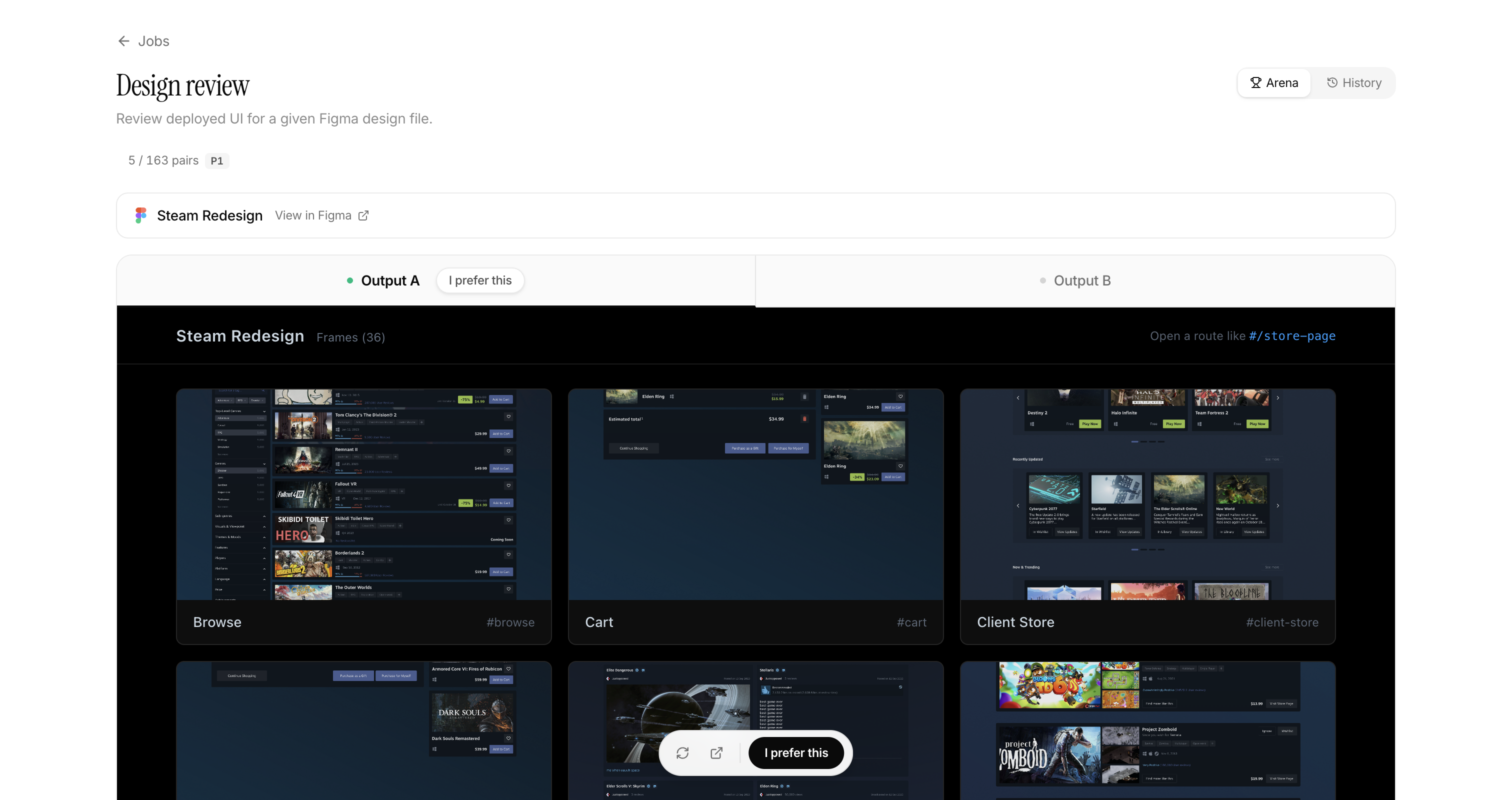}
  \caption{Preference arena interface for pairwise human evaluation. Two agent-built UIs are shown side-by-side for the same Figma design task, with frame-level thumbnails and deployed URLs. Position is randomized; agent identity is hidden.}
  \label{fig:preference_arena}
\end{figure}

Figure~\ref{fig:preference_arena} shows the pairwise preference evaluation interface used for human baselining. Annotators see two agent-built outputs side-by-side for the same Figma task, with deployed UI frames rendered at full fidelity. The original Figma design is linked for reference. Position (left/right) is randomized per vote to mitigate ordering effects. Annotators select ``I prefer this'' for the better output, with no access to agent identity or LLM scores.

\section{Task-Level Human Baseline}
\label{app:task_baseline}

Table~\ref{tab:task_pass_fail} reports per-task pass/fail classification from human expert evaluation, applying the quality threshold defined in Section~\ref{sec:results} ($\leq$3 = fail, $\geq$4 = pass). Tasks are sorted by overall pass rate to illustrate the difficulty gradient produced by the complexity stratification described in Section~\ref{sec:complexity_axes}.

\begin{table}[h]
  \caption{Per-task pass/fail from human expert evaluation (Figma-to-code, $n{=}31$ rated tasks). P/T = pass count / total rated runs for that harness on that task. Tasks sorted by overall pass rate.}
  \label{tab:task_pass_fail}
  \centering
  \small
  \begin{tabular}{lccccr}
    \toprule
    \textbf{Task ID} & \textbf{Claude} & \textbf{Codex} & \textbf{Gemini} & \textbf{Overall} & \textbf{Pass \%} \\
    \midrule
    \texttt{010ac575} & 0/3 & -- & 0/1 & 0/4 & 0\% \\
    \texttt{49bdd49a} & 0/2 & 0/1 & -- & 0/3 & 0\% \\
    \texttt{aafbd754} & 0/1 & 0/2 & -- & 0/3 & 0\% \\
    \texttt{d8611910} & 0/1 & -- & -- & 0/1 & 0\% \\
    \texttt{6d91b0b6} & 0/1 & 0/2 & 1/2 & 1/5 & 20\% \\
    \texttt{91ce9b15} & 0/2 & 0/2 & 1/1 & 1/5 & 20\% \\
    \texttt{20ef0a04} & 0/2 & 2/4 & -- & 2/6 & 33\% \\
    \texttt{7cd22b0d} & 0/1 & 0/1 & 1/1 & 1/3 & 33\% \\
    \texttt{85611489} & 1/1 & 0/2 & -- & 1/3 & 33\% \\
    \texttt{a93ffbb6} & 1/2 & -- & 0/1 & 1/3 & 33\% \\
    \texttt{792105af} & 2/2 & 0/3 & -- & 2/5 & 40\% \\
    \texttt{8b3bf60b} & 2/3 & 0/2 & -- & 2/5 & 40\% \\
    \texttt{63521424} & -- & 3/5 & 0/1 & 3/6 & 50\% \\
    \texttt{8efa99ee} & 1/1 & -- & 0/1 & 1/2 & 50\% \\
    \texttt{a713118e} & -- & 1/6 & 4/4 & 5/10 & 50\% \\
    \texttt{cead04c7} & 4/4 & 2/6 & -- & 6/10 & 60\% \\
    \texttt{3a8534f5} & 1/2 & 2/2 & 1/2 & 4/6 & 67\% \\
    \texttt{706c87ae} & 1/1 & 1/2 & -- & 2/3 & 67\% \\
    \texttt{7c46867b} & 0/1 & 2/2 & -- & 2/3 & 67\% \\
    \texttt{d0b0a0e3} & 2/2 & 0/1 & -- & 2/3 & 67\% \\
    \texttt{e03bd339} & -- & 1/1 & 1/2 & 2/3 & 67\% \\
    \texttt{f35388e3} & 0/1 & 2/2 & -- & 2/3 & 67\% \\
    \texttt{6af7bf85} & 2/2 & 2/2 & 1/3 & 5/7 & 71\% \\
    \texttt{43dd3b2d} & 2/2 & 3/3 & 0/1 & 5/6 & 83\% \\
    \texttt{c413f4df} & 0/1 & 5/5 & -- & 5/6 & 83\% \\
    \texttt{65829f23} & 0/1 & 6/6 & -- & 6/7 & 86\% \\
    \texttt{518a5ddc} & 1/1 & 2/2 & -- & 3/3 & 100\% \\
    \texttt{70278991} & 1/1 & 5/5 & -- & 6/6 & 100\% \\
    \texttt{9674e37a} & 1/1 & 6/6 & -- & 7/7 & 100\% \\
    \texttt{d00f7a9b} & -- & 9/9 & 1/1 & 10/10 & 100\% \\
    \texttt{d399b2f6} & 2/2 & -- & -- & 2/2 & 100\% \\
    \bottomrule
  \end{tabular}
\end{table}

\section{Experiment Versioning Infrastructure}
\label{app:versioning}

To support controlled ablations, each execution records a \texttt{versions} dictionary (e.g., \texttt{\{"skill": "v0"\}}) stored as a JSONB column alongside run metadata. \texttt{SKILL.md} files are versioned in a \texttt{versions/} subdirectory alongside the base skill file; at execution time, the agent runner overwrites the base \texttt{SKILL.md} with the specified version variant before launching the agent session. Analytics endpoints group results by \texttt{(harness, model, skill\_version)}, enabling per-condition comparison at both the task and aggregate level. The infrastructure supports additional ablation dimensions (e.g., prompt version, tool access) via the same \texttt{versions} dictionary without schema changes.

\end{document}